\documentclass[11pt]{article}

\usepackage[final]{acl}

\usepackage{times}
\usepackage{latexsym}
\usepackage{booktabs}
\usepackage[T1]{fontenc}

\usepackage[utf8]{inputenc}

\usepackage{microtype}

\usepackage{inconsolata}

\usepackage{graphicx}
\usepackage{algorithm}
\usepackage{algorithmic}
\definecolor{springgreen}{rgb}{0,0.78,0.34}
\title{UnAC: Adaptive Visual Prompting with Abstraction and Stepwise Checking for Complex Multimodal Reasoning}

\author{
 \textbf{Yifan Wang\textsuperscript{1}},
 \textbf{Yun Fu\textsuperscript{1}}
\\
\\
 \textsuperscript{1}Northeastern University
}

\begin{document}
\maketitle
\begin{abstract}
Although recent LMMs have become much stronger at visual perception, they remain unreliable on problems that require multi-step reasoning over visual evidence. In this paper, we present UnAC (Understanding, Abstracting, and Checking), a multimodal prompting method that strengthens reasoning for complex multimodal tasks in LMMs (e.g., GPT-4o, Gemini 1.5, and GPT-4V). To improve image understanding and capture fine details, we propose an adaptive visual prompting strategy that enables LMMs to focus on salient regions. We further design an image-abstraction prompt to effectively extract key information from images. In addition, we introduce a gradual self-checking scheme that improves reasoning by verifying each decomposed subquestion and its answer. Extensive experiments on three public benchmarks—MathVista, MM-Vet, and MMMU.
\end{abstract}

\section{Introduction}
In recent years, large language models (LLMs) have advanced significantly~\cite{brown2020language,achiam2023gpt,touvron2023llama,bubeck2023sparks,chowdhery2023palm,zhang2022opt}. 
Models such as GPT-3~\cite{brown2020language}, PaLM~\cite{chowdhery2023palm}, and Llama~\cite{touvron2023llama}, followed by GPT-4~\cite{achiam2023gpt},  PaLM~2~\cite{anil2023palm} and Gemini~\cite{team2023gemini}, have driven numerous breakthroughs in both industry and academia. 
This progress has spurred rising interest in large multimodal models (LMMs), with many approaches building powerful systems on open-source frameworks~\cite{liu2024visual,wu2023visual,dai2024instructblip,zhu2023minigpt}. 
Recently, the releases of GPT-4V(ision) and Gemini~1.5~Flash~\cite{team2023gemini} have drawn substantial attention for their strong capabilities in image understanding; however, these models still struggle with complex multimodal reasoning tasks~\cite{lu2023mathvista,yue2023mmmu}.

Since prompting approaches designed to improve the reasoning ability of LLMs in purely linguistic contexts~\cite{yao2024tree,wei2022chain,yao2022react,miao2023selfcheck,zheng2023take} have made significant progress, one might hope to transfer them directly to vision. However, because LMMs cannot decompose an image as easily as they can decompose a sentence, applying language-only prompts to enhance reasoning in the visual context is often ineffective. In visual question answering, major failure cases typically stem from misinterpreting the image or imprecisely summarizing its information. Such omissions or misunderstandings are closely related to limited fine-grained perception capability~\cite{yang2023set}.

Visual prompting has been investigated in a range of multimodal tasks, particularly as a way to improve fine-grained visual perception. Many methods encode masks (e.g., points, boxes, lines) that are fused with input features or directly overlay cues on the original image. Most recently, Yang \textit{et al.} proposed building a visual prompting mechanism by partitioning the image into semantically meaningful regions and overlaying them to strengthen the grounding ability of GPT-4V. However, for complex questions that require multi-step evidence extraction and reasoning, image partition alone is often insufficient to produce substantial gains in final reasoning performance.

\begin{figure*}
    \centering
    \includegraphics[width=1.0\linewidth]{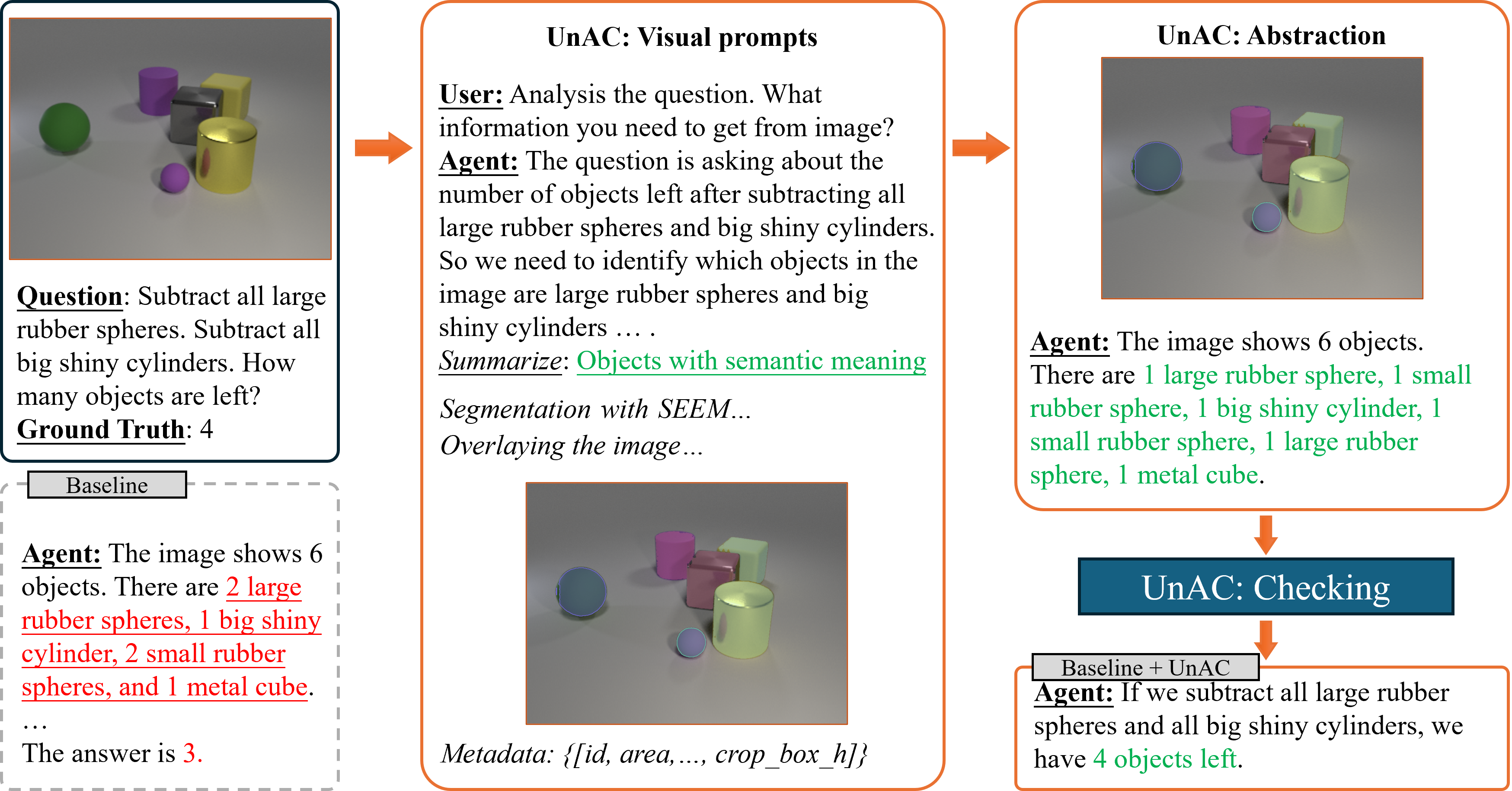}
    \caption{Example of using UnAC. In the original answer from the baseline method, the LMM incorrectly understands and describe the image which leads to the wrong answer. In UnAC which follows in the orange arrows, we first ask the LMM to analyze the question and answer what we need from the image. Then we can summarize the reply as the Objects with semantic meaning. Then employing SEEM to segment and overlay the image as visual prompts. Then abstracting the information of the image where with the markers, the LMM can correctly describe the image and abstract the right contexts. Finally, after the checking stage, we can get the right answer.}
    \label{frame}
\end{figure*}

In this work, we introduce \textbf{UnAC}—short for \textbf{Un}derstanding, \textbf{A}bstracting, and \textbf{C}hecking—a multimodal prompting framework designed to enhance complex reasoning in large multimodal models (LMMs).
UnAC follows a three-step prompting pipeline. First, we introduce an adaptive visual prompting scheme that places image-conditioned markers to direct the model’s attention toward specific regions, thereby reducing misunderstandings and missed details. By inspecting the image part by part, the model discovers more fine details and forms a more accurate global understanding. Second, to solve problems that require complex reasoning, we abstract the image into language based on the question. Inspired by how humans process visual information given a query, we extract key evidence both locally and globally by identifying question-relevant regions and converting them into textual descriptions using the established visual prompts. Third, because LMMs are prone to making errors at intermediate steps and global self-checking is often ineffective, we introduce a \emph{gradual self-checking} scheme that verifies each decomposed subquestion and its answer individually within the visual context.

We evaluate UnAC on three datasets that assess complex problem solving in the visual domain: MathVista~\cite{lu2023mathvista}, MM-Vet~\cite{yu2023mm}, and MMMU~\cite{yue2023mmmu}. To demonstrate generalization, we conduct experiments on two categories of LMMs: (a) powerful large-scale models, including GPT-4V and Gemini-1.5-Flash; and (b) relatively lightweight models, including LLaVA-v1.6-7B/13B. UnAC yields improvements across all models and datasets, indicating that our method is model-agnostic. Notably, on MathVista with Gemini-1.5-Flash, our method achieves a $6.4\%$ absolute gain.


To summarize, our main contributions are:
\begin{itemize}
    \item We introduce \textbf{UnAC} (\textbf{Un}derstanding, \textbf{A}bstracting, and \textbf{C}hecking), a simple yet effective multimodal prompting framework that strengthens complex multimodal reasoning in LMMs.
    \item We design an \emph{adaptive visual prompting} strategy that focuses the model on salient regions and reduces missed details, and couple it with question-conditioned \emph{image abstraction} and a \emph{gradual self-checking} procedure, yielding finer perception and more reliable step-wise reasoning.
    \item We validate UnAC on three benchmarks—MathVista, MM-Vet, and MMMU—achieving consistent improvements in complex visual reasoning across diverse LMMs.
\end{itemize}

\section{Related Work}
\paragraph{Prompting in LLMs.} We have observed significant advancements in large language models (LLMs) \cite{zhang2022opt, zhang2023llama, touvron2023llama, team2023gemini,brown2020language}. 
Although the size of LLMs has increased substantially, evoking their reasoning capabilities is still necessary with the use of more complicated designed queries, or prompting. 
Recently, various works have explored prompt engineering to enhance LLM capabilities. In-context learning has emerged as a widely used way to guide LLMs by providing a small set of task-specific examples. \cite{brown2020language, dong2022survey}. 
Building on this, techniques such as chain-of-thought and tree-of-thought \cite{wei2022chain, yao2024tree} have been introduced to improve performance in arithmetic, commonsense, and symbolic reasoning tasks. 
Most recently, Zheng \textit{et al.} \cite{zheng2023take} introduced Step-Back Prompting, which improves information retrieval by encouraging the model to reason through a more abstract formulation of the question. Miao \textit{et al.} \cite{miao2023selfcheck} introduced a general-purpose zero-shot verification schema for recognizing errors made in the reasoning process of math problems. 
However, their methods highly rely on that the language is easy to be decomposed. It is hard to be generalized to the question in the visual context where images are hard to decompose. 
\paragraph{Prompting in LMMs} 
Recent prompting research in LLMs has shown that model behavior can be substantially improved without parameter updates, especially for reasoning-intensive tasks\cite{wang2023seggpt,zou2024segment,kirillov2023segment, chen2022focalclick,shtedritski2023does}. Existing visual prompting methods mainly fall into two families. One line of work injects prompt signals into latent representations \cite{zou2024segment,kirillov2023segment}, while another directly modifies the image with explicit visual markers such as circles, boxes\cite{shtedritski2023does}, or highlighted regions \cite{yang2023set}. 
Although prior studies suggest that pixel-level visual prompts can improve perception, most of them focus on highlighting only a small number of visually salient regions or objects. 
Research on prompting large multimodal models remains relatively limited, partly because many earlier open-source LMMs did not yet have sufficient capability to fully benefit from such complex prompting schemes.
Recently, GPT-4V was released, accompanied by a comprehensive qualitative study \cite{yang2023dawn}. The authors in \cite{yang2023dawn} employed a similar prompting strategy as RedCircle \cite{shtedritski2023does} to prompt GPT-4V. 
More recently, Yang \textit{et al.} \cite{yang2023set} introduced a method that decomposes an image into semantically meaningful regions and overlays these regions as visual cues to improve GPT-4V’s grounding capability.
CCoT \cite{mitra2023compositional} is designed as a zero-shot CoT prompting method to extract compositional knowledge from an LMM with utilizing scene graphs. However, both of these works can not solve the problem based on the abstract images such as geometry problem solving and math word problems.

\section{UnAC: Understanding, Abstracting, and Checking}
When people solve difficult visual problems, they usually proceed in stages: they first establish a global understanding, then isolate the evidence relevant to the question, and finally verify the reasoning before answering.
Moreover, for a complicated question, we usually need a second look at the reasoning process and check it with the image to avoid some simple mistakes.
Inspired by this common sense, we propose UnAC which means understanding, abstracting, and checking for synergizing the complicated reasoning in the visual context of large multimodal models.

\subsection{Adaptive Visual Prompts.}
Precisely capturing the details in the image is not straightforward for LMMs. It is hard to correct the misunderstanding of the image by itself because decomposing the image is not easy. Since LMMs are developed based on the LLMs, their abilities of language reasoning are much better than visual reasoning. It means that LMMs can perform better on analyzing the problem than analyzing the image. 
Therefore,  we propose to build effective and adaptive multimodal prompts based on the analysis of the question.
Asking the model to analyze the question and find what information we need to get from the image. We conclude the response into two kinds: Objects with semantic meaning and symbols with literal meaning. For objects with semantic meaning, we employ segmentation models to automatically segment the image.
For symbols with literal meaning, we use optical character recognition (OCR) methods to detect the texts. Based on the metadata, we first denoising regions based on the stability score output by the segmentation/OCR methods. 

In the Figure. \ref{frame}, we show a successful case. For this question of subtracting the items, it requires LMMs to correctly recognizing each item in the picture which is related to objects with semantic meanings. Accordingly, we construct the visual prompts through image segmentation, so that the LMM can form a more accurate understanding of the visual content.

\begin{figure}
\includegraphics[width=1\linewidth]{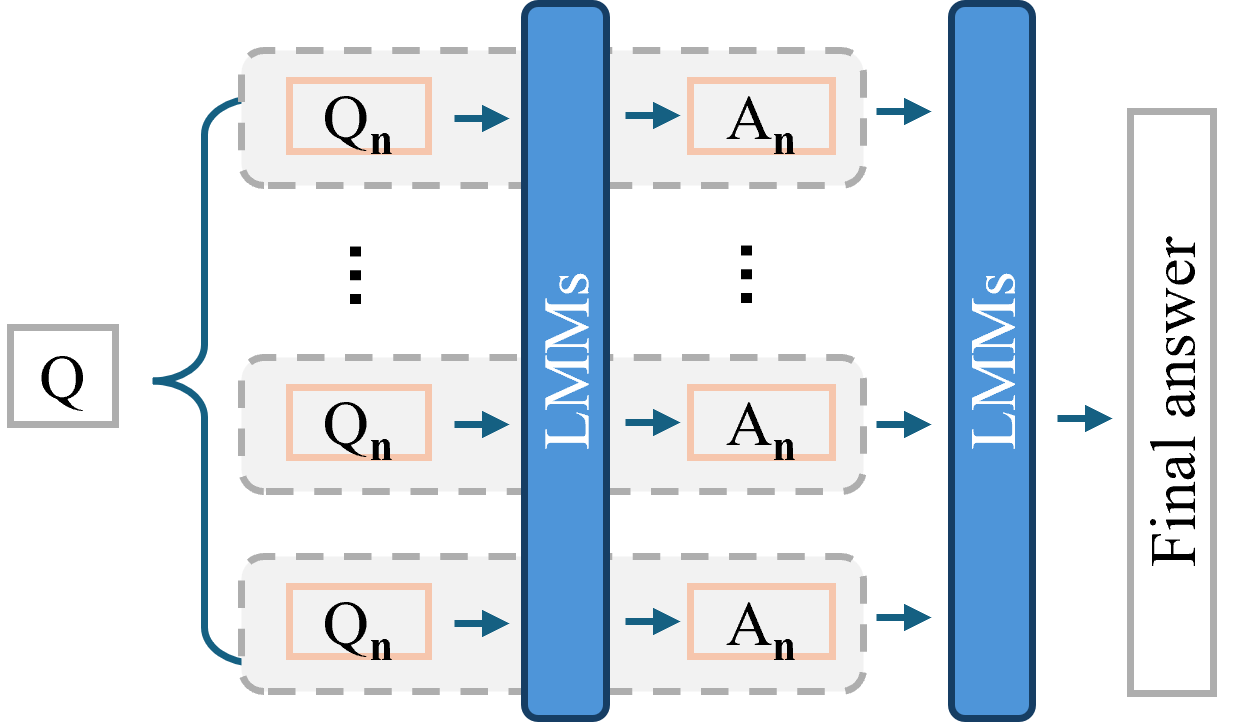}
\caption{Illustration of the gradual-checking process. The model first breaks the problem into sub-questions, answers them sequentially, and then revisits these intermediate results before producing the final answer. }
\label{osc}
\end{figure}

\subsection{Image Abstraction} 
The visual prompts can make a better understanding of the image since the markers can catch more attentions on some local information. 
Partitioning the image makes it decomposable when LMMs understand the image. However, only visual prompts have limited improvements for solving complicated problems. 
Except for understanding the image, LMMs need to correctly abstract the image to filter the useless information to solve the problem.
Without prompts of abstraction, the reasoning might be misdirected due to the markers in the image. 
Therefore, to fully utilize the visual prompts and get better reasoning, we need to abstract the information which is the most related to the question. 
Firstly, we ask LMMs to describe the picture to abstract the global information. Then based on the analysis of the question and the prompts, we ask LMMs to find the most related regions to get more details based on the markers in the image.

\subsection{Gradual checking}

Moreover, for some complicated questions, we usually need a second look at the image with the reasoning progressing. As discussed in \cite{ling2024deductive}, checking the whole reasoning process is usually ineffective for LLMs and our experiments show similar results in LMMs. 
However, to correct the mistake made in one step is more effective.
When verifying intermediate reasoning steps, an important consideration is that each step can only be judged correctly in relation to its surrounding context.
For a question in words, the context includes the question and previous steps only. So the checking is largely dependent on the accuracy of the previous steps which is highly unstable. 
In the visual question answering, the information from the image becomes extra contexts which are important references for self-checking. It can be more reliable when LMMs have a good understanding of the image. 

Then, we design a gradual checking prompting for better reasoning. 
Firstly, we let LMMs decompose the question into multi sub-questions $ [ Q_0, Q_1, \dots, Q_n ]$ and give the answer of each sub-questions. The answers are denoted as $[A_0, A_1, \dots, A_n]$. 
In the checking stage, we check gradually. When checking $Q_i$ and $A_i$, we refer the context of the previous questions and checked answers $ [ Q_0, Q_1, \dots, Q_i ]$ and  $[A^{\prime}_0, A^{\prime}_1, \dots, A^{\prime}_i]$. In the last step of checking, LMMs will infer the final answer based on all questions and answers.

\section{Experiments}
\label{gen_inst}
\subsection{Setup}
\paragraph{Tasks and datasets.}
We experiment with the following two tasks that need complicated reasoning: (a) Mathematical reasoning in the visual context, and (b) Complicated VQA.
{\it Mathematical reasoning:} We evaluate MathVista \cite{lu2023mathvista} for this task. MathVista is a benchmark for mathematical reasoning in visual settings, designed to evaluate how well foundation models solve math problems grounded in images.  It contains various kinds of sub-tasks to evaluate the model's visual understanding of mathematical problems solving in different perspectives of reasoning skills. 
{\it Complicated VQA:} For this task, we evaluate two datasets called: 
MM-Vet \cite{yu2023mm} and MMMU \cite{yue2023mmmu} respectively. 
MM-Vet is designed to assess large multimodal models on challenging vision-language tasks that require the integration of six core capabilities: Recognition, Knowledge, Optical Character Recognition (OCR), Spatial Awareness, Language Generation, and Math. MMMU evaluates advanced multimodal understanding and reasoning in domain-specific settings, requiring models to solve expert-level tasks.
\begin{table*}[t]

    \centering
\caption{Performance comparison on the {\it testmini} split of MathVista (\cite{lu2023mathvista}. 
The table reports overall accuracy as well as a detailed breakdown across representative categories of visual mathematical reasoning, allowing us to examine not only the aggregate gains of UnAC but also where the improvements are most pronounced. The results show that UnAC consistently improves performance over the corresponding baselines across different model families and evaluation dimensions.
 }
\vspace{2mm}
\resizebox{1.0\textwidth}{!}
   {
   \begin{tabular}{lc|ccccc|ccccccc}
   \toprule
    {Method} & {ALL}  & FQA & GPS & MWP & TQA & VQA & ALG & ARI & GEO & LOG & NUM & SCI & STA\\
    \midrule
    \multicolumn{14}{c}{\textit{Human performance}}\\
    \midrule
    Human Performance  & $60.3$ & $59.7$ & $48.4$ & $73.0$ & $63.2$ & $55.9$ & $50.9$ & $59.2$ & $51.4$ & $40.7$ & $53.8$ & $64.9$ & $63.9$ \\
    \midrule
    \multicolumn{14}{c}{\textit{Heuristics baselines}}\\
    \midrule
Random chance&$17.9$&$18.2$&$21.6$&$3.8$&$19.6$&$26.3$&$21.7$&$14.7$&$20.1$&$13.5$&$8.3$&$17.2$&$16.3$\\
Frequent guess&$26.3$&$22.7$&$34.1$&$20.4$&$31.0$&$24.6$&$33.1$&$18.7$&$31.4$&$24.3$&$19.4$&$32.0$&$20.9$\\
\midrule
\multicolumn{14}{c}{\textit{Closed-sourced Large Multimodal Models (LMMs)
}}\\
   \midrule
   Gemini-1.0-pro-vision  & $41.0$&$36.4$&$36.5$&$43.0$&$57.5$&$36.3$&$39.8$&$37.6$&$38.0$&$10.8$&$29.8$&$52.4$&$45.5$\\
      Gemini-1.0-pro-vision + UnAC & $\mathbf{47.4}\textcolor{springgreen}{ \scalebox{0.75}{${(+6.4)}$}}$&${49.4}$&$39.5 $&$45.2$&$62.7$&$42.5$ &$43.8$&$44.2$&$40.1$&${29.7}$&${36.1}$&$54.9$&$57.1$   \\
          \midrule
          
        Gemini-1.5-flash &$53.2$&$51.6$&$56.8$&$52.1$&$67.5$&$40.1$ &$	59.4$&$44.0$&$52.7$&$25.3 $&$36.4 $&$60.8$&$57.5$ \\
        
          Gemini-1.5-flash  + UnAC& $\mathbf{56.6}\textcolor{springgreen}{ \scalebox{0.75}{${(+ 3.4)}$}}$ & ${57.3}$&${59.3}$&$54.1$&${71.6}$&${41.4}$&${62.8}$&${46.6}$&${59.5}$&${30.9}$&${37.7}$&${65.3}$&${65.8}$  \\
          \midrule
          GPT-4V & $50.7	$&$43.6$&$	50.5$&$	{57.5}$&$	65.2	$&$38.4$&$	53.0$&$49.0$&$	51.0	$&$21.6	$&$20.1	$&$63.1$&$	55.8$ \\
           GPT-4V +SoM \cite{yang2023set}&$51.2\textcolor{springgreen}{ \scalebox{0.75}{${(+ 0.5)}$}}$&$50.5$&$52.9$&$49.7$&$64.8$&$37.2$ &$53.4$&$44.0$&$51.2$&$18.9 $&$32.4 $&$62.8$&$57.5$ \\
        GPT-4V + CCoT \cite{mitra2023compositional}& $51.8\textcolor{springgreen}{ \scalebox{0.75}{${(+ 1.1)}$}}	$&$46.2$&$	50.2$&$	{58.2}$&$	64.2	$&$40.4$&$	55.0$&$48.2$&$	51.2	$&$21.6	$&$20.1	$&$57.1$&$	59.2$\\
        GPT-4V + SKETCHPAD \cite{hu2024visual}& $52.0\textcolor{springgreen}{ \scalebox{0.75}{${(+ 1.3)}$}}	$&$44.2$&$	52.6$&$	{60.5}$&66.4 & 37.8&53.5& 48.2& 51.9& 21.2& 19.8& 63.8& 55.6\\
          GPT-4V  + UnAC& $\mathbf{57.6}\textcolor{springgreen}{ \scalebox{0.75}{${(+ 4.9)}$}}$&$47.3$&${61.1}$&$55.2$&${69.7}$&${48.9}$&${60.9}$&${50.1}$&${58.5}$&$18.9$&$35.4$&${60.7}$&${57.8}$  \\
          \midrule
      \multicolumn{14}{c}{\textit{Open-sourced Large Multimodal Models (LMMs)
}}\\
\midrule
        
          LLaVA-OneVision-7B&$34.8$&$43.6$&$21.6$&$27.9$&$43.4$&$37.8$&$26.5$&$32.0$&$23.3$&$19.9$&$24.9$&$49.1$&$44.6$ \\
           LLaVA-OneVision-7B + SoM&$34.6\textcolor{red}{ \scalebox{0.75}{${(- 0.2)}$}}$&$43.7$&$19.6$&$29.6$&$43.2$&$37.1$&$26.9$&$31.6$&$20.8$&$19.6$&$25.2$&$50.2$&$43.9$ \\ 
         LLaVA-OneVision-7B + UnAC& $\mathbf{36.2}\textcolor{springgreen}{ \scalebox{0.75}{${(+ 1.4)}$}}$ & $49.5$&$20.1$&$28.5$&$44.4$&$38.8$& $32.5$&$31.2$&$21.1$&$18.4$&$25.1$&$50.3$&$44.6$ \\
         \midrule
         LLaVA-v1.6-13B & $35.8$&$45.3$&$21.6$&$29.5$&$43.0$&$37.9$&$24.9$&$33.9$&$23.8$&$13.5$&$27.7$&$49.1$&$48.1$  \\
          LLaVA-v1.6-13B  + UnAC & $\mathbf{37.8}\textcolor{springgreen}{ \scalebox{0.75}{${(+ 2.0)}$}}$&$37.5$&$31.7$&$30.7$&$53.8$&$38.5$&$33.4$&$34.6$&$32.2$&$10.8$&$25.7$&$53.3$&$44.9$  \\
          
          \midrule
          InternVL2.0-8B & 67.3&	72.5&	73.6&	69.9	&66.5&	50.3&	70.1	&57.5&	71.5	&27.0	&43.1&	65.6&	79.1 \\
        InternVL2.0-8B + SoM \cite{yang2023set} & 67.2 $\textcolor{red}{ \scalebox{0.75}{${(- 0.1)}$}}$&	75.0&	72.4&	72.0	& 65.2&	51.3&	73.2	&58.5&	72.5	&22.5	&41.0&	65.6&	79.1 \\
        InternVL2.0-8B + CCoT \cite{mitra2023compositional}& 68.2 $\textcolor{springgreen}{ \scalebox{0.75}{${(+ 0.9)}$}}$&75.6&76.2&72.3&65.8&49.1& 69.5& 60.5&	70.0	&25.2& 42.5& 68.6&	75.2\\
        InternVL2.0-8B + SKETCHPAD \cite{hu2024visual} &69.2$\textcolor{springgreen}{ \scalebox{0.75}{${(+ 1.9)}$}}$&77.2&77.6&72.8&70.1&48.3& 73.2& 62.1& 76.2& 28.3& 43.2& 64.6&79.3\\
          InternVL2.0-8B  + UnAC& $\mathbf{71.6}\textcolor{springgreen}{ \scalebox{0.75}{${(+ 4.3)}$}}$&77.3&79.6&75.0&72.1&54.0 & 75.4 & 62.5&	75.5&31.0	&44.8&	67.2&	80.7 \\
      
      \bottomrule
    
    \end{tabular}}
    \label{mathviste}
\end{table*}

\paragraph{Models.}
To evaluate the generality of UnAC, we experiment with a diverse set of state-of-the-art LMMs, including the closed-source models GPT-4V and Gemini-1.5-Flash,
relatively small LMMs including LLaVA-v1.6-7B/13B \cite{liu2024visual}, LLaVA-OneVision \cite{li2024llava} and internVL2.0-8B \cite{chen2024internvl}.  For the closed-source LMMs, we utilize the official API to make the evaluation. We use 'gpt4-turbo' and 'gemini-1.5-flash' for GPT4-V and Gemini respectively. For the open-source models, we evaluate in a single RTX 6000. We set the temperature to $0.0$ for all LMMs. We use SEEM \cite{zou2024segment} for segmentation and easyOCR for building the visual prompts. 
Moreover, we also compare with two chain-of-thought methods including CCoT \cite{mitra2024compositional} and SKETCHPAD \cite{hu2024visual} to show the superiority of UnAC as a training-free method.
\paragraph{Evaluation.} Each benchmark provides a single ground-truth answer for every question, where the target may take the form of a number, a word, a short phrase, or a multiple-choice option. We use accuracy (ACC) as the primary evaluation metric. Because LMM outputs are often free-form and may not exactly match the reference string, we follow prior works \cite{lu2023mathvista, yu2023mm}, and use a few-shot GPT-4-based answer matching procedure to judge whether a prediction is semantically equivalent to the ground-truth answer.

\subsection{Results}

\paragraph{Mathematical reasoning in the visual context.}
In Table \ref{mathviste}, we present results on the MathVista benchmark \cite{lu2023mathvista}, where UnAC consistently improves performances across all models. Specifically, we achieve a $4.9\%$ gain on GPT-4V and $3.4\%$ on Gemini-1.5-flash. For LLaVA-v1.6-7B/13B, the improvements are $2.6\%$ and $2.0\%$, and for LLaVA-OneVision-7B, we observe a $1.4\%$ gain—outperforming SoM \cite{yang2023set} on the same model. Notably, our method boosts InternVL2.0-8B by $4.3\%$, significantly surpassing chain-of-thought approaches like CCoT ($0.9\%$) and SKETCHPAD ($0.8\%$), which struggle to improve strong baselines.

Across sub-tasks, our method shows marked gains on the most challenging ones: a $8.6\%$ improvement on Geometry Problem Solving (GPS) with GPT-4V and $4.1\%$ on TQA with Gemini. These tasks require complex, multi-step reasoning, which benefits from better visual abstraction and our self-checking scheme. For simpler tasks like VQA and FQA, the gains confirm the effectiveness of our adaptive visual prompting.
Overall, the consistent improvements demonstrate that UnAC is a model-agnostic prompting strategy. However, stronger models like GPT-4V and InternVL benefit more, as the effectiveness of both visual prompting and self-checking depends on the model’s reasoning capability—further discussed in Sec. 4.3.

\begin{table}[t]
    \centering
 \caption{ Accuracy scores on the MM-Vet and the validation set of MMMU.}
 \label{mmvet}
 \resizebox{0.45\textwidth}{!}
   {
    \begin{tabular}{l| c c}
    \toprule
    Method & MM-Vet &MMMU\\
    \midrule
        LLaVA-v1.6-7B & ${47.5}$ & $36.9$  \\
       LLaVA-v1.6-7B + Ours & $\mathbf{48.5}\textcolor{springgreen}{ \scalebox{0.75}{${(+ 1.0)}$}}$ & $\mathbf{37.4}\textcolor{springgreen}{ \scalebox{0.75}{${(+ 0.5)}$}}$   \\
       \midrule
       LLaVA-OneVision-7B & $57.5$ & $48.8$  \\
       LLaVA-OneVision-7B + Ours & $\mathbf{60.2}\textcolor{springgreen}{ \scalebox{0.75}{${(+ 2.7)}$}}$ & $\mathbf{51.0}\textcolor{springgreen}{ \scalebox{0.75}{${(+ 2.2)}$}}$   \\
       \midrule
       Gemini-1.5-flash & ${62.2}$ & $56.1$  \\
       Gemini-1.5-flash + Ours & $\mathbf{64.9}\textcolor{springgreen}{ \scalebox{0.75}{${(+ 2.7)}$}}$ & $\mathbf{60.9}\textcolor{springgreen}{ \scalebox{0.75}{${(+ 4.8)}$}}$   \\
       \midrule
        InternVL2.0-8B & $60.0$ & $51.8$  \\
       InternVL2.0-8B + UnAC & $\mathbf{63.3}\textcolor{springgreen}{ \scalebox{0.75}{${(+ 3.3)}$}}$ & $\mathbf{54.7}\textcolor{springgreen}{ \scalebox{0.75}{${(+ 2.9)}$}}$   \\
       \midrule
       GPT4-V & $67.2$ & $57.2$  \\
        GPT4-V + SoM \cite{yang2023set} & $66.0 \textcolor{red}{ \scalebox{0.75}{$(- 1.2)$}}$& $57.2$  \\
       GPT4-V + CCoT & $67.7\textcolor{springgreen}{ \scalebox{0.75}{${(+ 0.5)}$}}$  & $58.7$ \textcolor{springgreen}{ \scalebox{0.75}{${(+ 1.5)}$}}  \\
       GPT4-V + SKETCHPAD  & ${69.3}\textcolor{springgreen}{ \scalebox{0.75}{${(+ 2.1)}$}}$ & ${59.7}\textcolor{springgreen}{ \scalebox{0.75}{${(+ 2.5)}$}}$  \\
       GPT4-V + Ours & $\mathbf{70.3}\textcolor{springgreen}{ \scalebox{0.75}{${(+ 3.1)}$}}$ & $\mathbf{60.7}\textcolor{springgreen}{ \scalebox{0.75}{${(+ 3.5)}$}}$   \\

       \bottomrule

    \end{tabular}}
\end{table}
\paragraph{Complicated VQA.} 
In Table \ref{mmvet}, we show the results on the MM-Vet \cite{yu2023mm} and MMMU \cite{yue2023mmmu}. In these two datasets, the questions are more generalized with a relatively simple reasoning process.  Our method still makes improvements on all models. We make an improvement of $3.1\%$ on GPT-4V with our method  and make the largest increase of $4.8\%$ on Gemini-1.5-flash on MMMU. Compared to the chain-of-thought methods, UnAC performs better. Also, it indicates the necessity  of self-checking that applying SoM \cite{yang2023set} on GPT-4V is harmful to answer the complicated question.
For LLaVA-OneVision-7B, we achieve the improvements of  $2.7\%$ on MM-Vet.

The gap between the increase on Gemini/GPT4-V and the increase of LLaVA-v1.6-7B is larger compared to that on MathVista. In these two datasets, they require more comprehensive vision-language capabilities and abundant knowledge reserve on various topics. Therefore, in those two datasets, understanding can be more important than abstracting and reasoning.  

\begin{figure*}[t]
\includegraphics[width=0.48\linewidth]{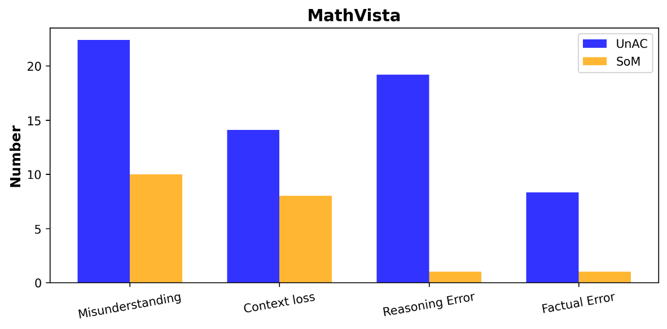}
\includegraphics[width=0.48\linewidth]{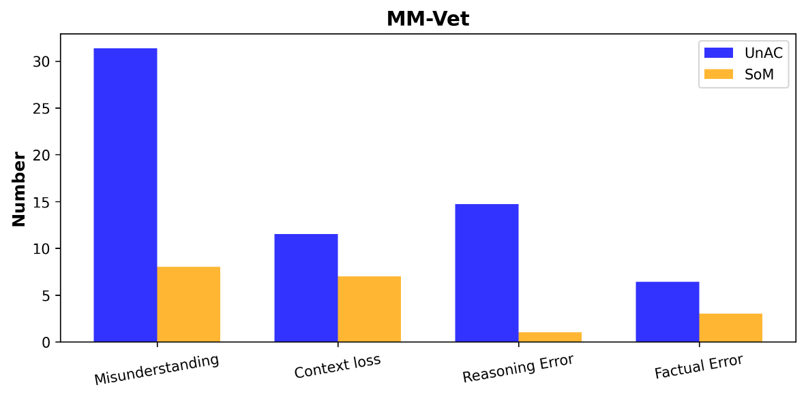}

\caption{Corrected error analysis and comparison of UnAC and SoM. The left plot shows the comparison on MathVista while the right one represents that on MM-Vet: four classes of errors are corrected by the UnAC or SoM. The baseline model is GPT-4V for both methods.}
\label{error}
\end{figure*}

\subsection{Analysis}
{\bf Corrected error analysis.}
To better understand why UnAC improves over the GPT-4V baseline, we analyze the cases in which our method changes the original prediction on MathVista and MM-Vet. On MathVista, UnAC corrects 25.4\% of the errors made by the baseline while introducing 5.5\% new errors. On MM-Vet, the method fixes 20.1\% of the baseline mistakes and introduces 6.2\% additional errors. These numbers suggest that the gains mainly come from recovering a meaningful portion of previously incorrect predictions, while the number of newly introduced errors remains relatively limited. To further examine the source of these corrections, we manually inspect the examples where UnAC succeeds and group them into four categories. 
{\bf Misunderstanding} refers to cases where the baseline fails because it interprets the image incorrectly, while UnAC produces a more accurate visual understanding. {\bf Context loss} includes cases where the baseline overlooks relevant visual evidence or omits useful details that are necessary for answering the question. {\bf Reasoning error} describes examples in which the retrieved evidence is relevant, but the model still fails to combine it correctly to reach the answer. {\bf Factual error} refers to failures caused by incorrect factual knowledge rather than by purely visual or reasoning-related mistakes. This categorization helps us separate gains from improved perception and context extraction from those arising from stronger reasoning or better factual grounding.
\begin{table*}[t]
    \centering
\caption{
Ablation results for different visual prompting designs in UnAC on the MathVista {\it testmini} split \cite{lu2023mathvista}, using Gemini-1.5-Flash as the backbone model. The comparison shows how different prompt constructions affect overall performance across representative reasoning categories.}
\vspace{2mm}
  \setlength{\tabcolsep}{7.2mm}
\resizebox{1.0\textwidth}{!}
   {
   \begin{tabular}{lc|ccccc}
   \toprule
    {Method} & {ALL}  & FQA & GPS & MWP & TQA & VQA \\
    \midrule
   Baseline &$53.2$&$51.6$&$56.8$&$52.1$&$67.5$&$40.1$   \\
  Segmentation only & $54.2$&$53.2$&$57.8$&$53. 0$&$68.4$&$39.6$  \\
       OCR only & $53.3$&$51.6$&$57.3$&$51.1$&$68.7$&$40.1$ \\
    Segmentation + OCR& $54.4$&$53.2$&$57.8$&$53. 0$&$68.4$&$39.6$ \\
       Adaptive visual prompts  & $\mathbf{56.6}$ & $\mathbf{57.3}$&$\mathbf{59.3}$&$\mathbf{54.1}$&$\mathbf{71.6}$&$\mathbf{41.4}$\\   
      \bottomrule
    
    \end{tabular}}
    \label{vpp}
\end{table*}

\begin{table*}[t]
    \centering
\caption{Accuracy scores using GPT-4V on the \textit{testmini} subset of MathVista \cite{lu2023mathvista} under different checking. For the global checking, we use a simple prompt of `Please check your answer if there are any errors.'}
\vspace{2mm}
  \setlength{\tabcolsep}{6.2mm}
\resizebox{1.0\textwidth}{!}
   {
   \begin{tabular}{lc|ccccc}
   \toprule
    {Method} & {ALL}  & FQA & GPS & MWP & TQA & VQA \\
    \midrule
    w/o Checking & $52.7$&$55.5$&$53.4$&$49.2$&$65.8$&$37.7$   \\
   Global Checking & $53.4$&$55.5$&$53.4$&$49.2$&$65.0$&$42.7$  \\
  Gradual Checking& $\mathbf{57.6}$&$47.3$&$61.1 $&$55.2$&$69.7$&$48.9$  \\

      \bottomrule
    \vspace{-2mm}
    \end{tabular}}
    \label{checking}
\end{table*}

\textit{MathVista.} As shown in Figure~\ref{error} (left), about $35\%$ of corrected errors stem from image misunderstanding, and $23\%$ from missing context. Together, roughly $58\%$ of errors are rectified by our adaptive visual prompts, which help LMMs better perceive image details. In contrast, some errors occur despite correct perception—due to flawed reasoning or missing factual knowledge. While SoM’s visual markers aid perception, they do little to improve reasoning, limiting its effectiveness in fixing such errors. Our self-checking scheme addresses this gap, accounting for $42\%$ of corrections through step-by-step validation.

\textit{MM-Vet.} In Figure~\ref{error} (right), $49\%$ of corrected errors are due to misunderstanding, and $18\%$ to lost visual context—both improved by UnAC. Reasoning and factual errors account for another $23\%$ and $10\%$, respectively. Since MM-Vet emphasizes fine-grained image understanding over complex reasoning, visual prompts play a larger role. As a result, both UnAC and SoM mainly correct perception-related errors. Still, UnAC demonstrates a notable $33\%$ improvement in reasoning-related cases, thanks to its self-checking mechanism.

\paragraph{Discussion.}Compared to the first two classes and the last two classes, the number of errors removed by correctly understanding the image and capturing more useful contexts is more than that removed by the accurate reasoning process.  It indicates that the reasoning step is still a bottleneck of how well UnAC can perform for tasks such as MathVista which requires more complex reasoning.

\begin{figure}[t]
\includegraphics[width=0.49\linewidth]{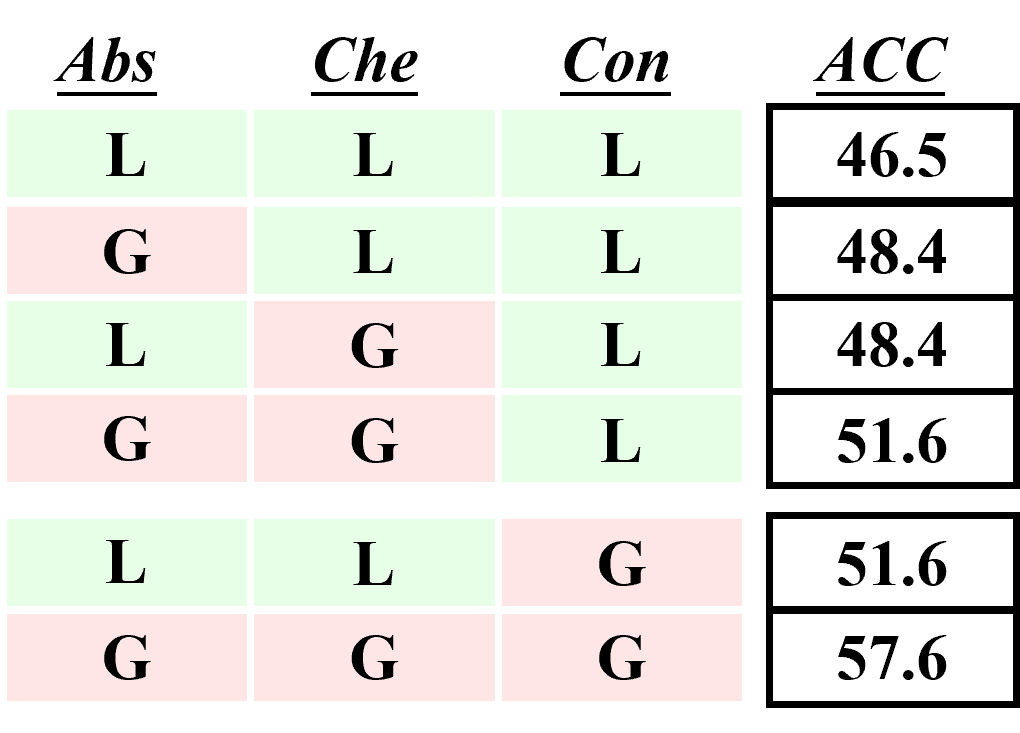}
\includegraphics[width=0.49\linewidth]{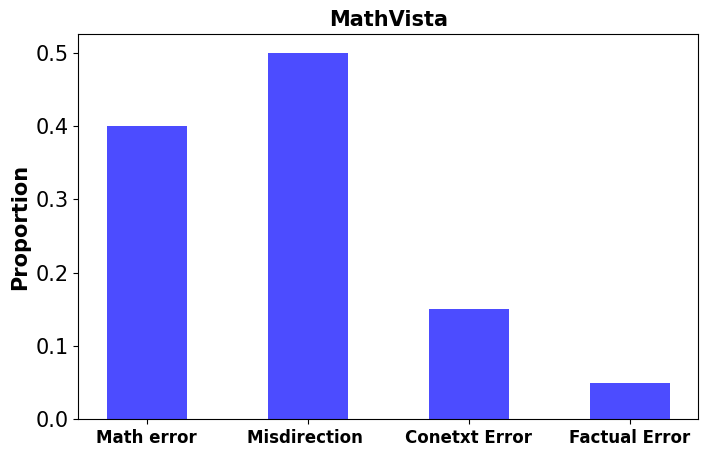}
\caption{\textbf{Left}: The overall accuracy of changing different part of UnAC on the textmini dataset of
MethVista \cite{lu2023mathvista}.  L means LLaVA-v1.6-7B and G means GPT-4V. \textit{Abs}, \textit{Che} and \textit{Con} represent the abstracting, checking and conclusion stages respectively. \textbf{Right}: The error analysis on MethVista with Gemini-1.5-flash using global checking.}
\label{temp}
\end{figure}

\paragraph{How the abstraction and self-checking affect the final answer?}
As we discussed in Sec 4.2, the improvements made by UnAC are influenced by the original capability of the baseline LMMs. Although it makes sense, we want to find out how it influences our method. We conduct experiments on changing the models which is used in abstracting, checking, and final reasoning mainly with LLaVA-v1.6-7B and GPT-4V.
As shown in Figure \ref{temp} (left), we replace the LLaVA-v1.6-7B with GPT-4V on different roles in our prompting process.
Comparing the first Four rows, the final conclusion performs much better when replacing LLaVA-v1.6-7B with GPT-4V for performing abstracting, and checking respectively. 
The best performance is contributed by using GPT-4V to make both abstracting and checking among these three ablations. It indicates that better abstracting and checking are helpful for increasing the overall performance. 
However, comparing the four rows and the bottom row, although GPT-4V may provide the accurate answer to the question in the checking stage, the LLaVA-v1.6-7B still infers bad reasoning in the last step. 
Moreover, comparing the fourth row and fifth row, we can find that even LLaVA-v1.6-7B provides the bad prompts, GPT-4V still has the ability of self-correction in conclusion.
Although improving the abstracting and checking can lead to better performance, the reasoning abilities of LMMs are still the bottleneck of how well UnAC can perform in solving complicated questions.

\paragraph{Why do visual prompts need to be adaptive?}
In this ablation, we want to show the effect of making the visual prompts adaptive. As shown in Table \ref{vpp}, we conduct experiments on applying different types of visual prompts. Comparing the first two lines, the improvements when employing the segmentation or OCR only are very limited.
Although partitions can help the LMMs to focus on a certain part of the image, they also increase the risk of focusing on the wrong regions on the image. 
Since the whole picture has been overlayed everywhere, it may confuse the attention of LLMs. 
Adding boxes on the image to let LMMs focus on certain parts, it also increase the risk of incorrect regions which are useless for the question answering.
Moreover, for some tasks, markers of segmentation or boxes from OCR is not helpful such as solving a geometry problem or understanding a function plot. Both prompts can not provide much useful information. 

\paragraph{Global checking or gradual checking.}
To prove that LMMs can not perform the global checking in an effective way like LLMs \cite{ling2024deductive}, we conduct experiments on comparing the performance of global checking prompting with the proposed one-step checking method. As shown in Table \ref{checking}, compared to UnAC without checking, the performance of using the global checking shows very limited improvement overall. Although it increase the accuracy of the textbook question answering and visual question answering tasks, it makes the qualities on tasks of math word problem and geometry problem solving worse. 
We also conduct experiments on analyzing the errors made by the global checking. As shown in Figure. \ref{temp} (right), we define another set of errors which are related to the reasoning process only. The classes of errors are 
(1) Math error: The additional mathematical errors like computation and mathematical inference; 
(2) Misdirection: Leading to focusing the wrong regions of the images.
(3) Context error: Incorrectly understanding the images or solutions in the previous steps.
Misdirection and Math errors are the most frequent errors occurring which have $50\%$ and $39\%$. It indicates that the global checking easily makes the reasoning process into the wrong direction due to the limitation of the reasoning ability of LMMs.

\section{Limitations}
Nevertheless, visual prompts are neither necessary nor possible to work in all scenarios. For instance, when facing highly abstract problems like geometry problem solving, the understanding of the image mostly depends on the original capability or the trained dataset of the LMMs since even a simple shape like a heptagon might be misidentified. 
How to effectively develop visual prompts for such problems is still a challenging topic and that's one of the future works we will target on.
\section{Conclusion}
In this paper, we propose a novel multimodal prompting method, namely UnAC (Understanding, Abstracting, and Checking), to synergize reasoning for complicated problems in visual context of LMMs. UnAC consists of an adaptive visual prompting building, the prompts of image abstraction and a gradual checking scheme. Suffecient experiments show the effectiveness of UnAC on improving the ability of complicated multimodal reasoning.

\bibliography{custom}

\appendix

\end{document}